\documentclass[11pt]{article}
\usepackage[letterpaper,margin=1in]{geometry}
\usepackage{times}
\usepackage[round]{natbib}
\usepackage{amsmath,amsfonts,bm}

\def\eqref#1{equation~\ref{#1}}
\def\1{\bm{1}}

\DeclareMathAlphabet{\mathsfit}{\encodingdefault}{\sfdefault}{m}{sl}
\SetMathAlphabet{\mathsfit}{bold}{\encodingdefault}{\sfdefault}{bx}{n}

\usepackage{hyperref}
\usepackage{url}
\usepackage{graphicx}
\usepackage{booktabs}
\usepackage{tabularx}
\usepackage{array}
\usepackage{float}
\newcolumntype{Y}{>{\raggedright\arraybackslash}X}
\usepackage{multirow} 
\usepackage{subcaption}
\usepackage{float}
\usepackage{placeins}
\usepackage{enumitem}
\usepackage{xcolor}

\begin{document}
\title{What the Trace Shows and What the Label Says: Auditing Behavioral Detectors in Agent Evaluation}

\author{
  Idunnuoluwa A. Adeniji$^{1}$, Shilpika$^{2}$, and Michael E. Papka$^{1,2}$\\[4pt]
  $^{1}$Department of Computer Science, University of Illinois Chicago, Chicago, IL, USA\\
 $^{2}$Argonne Leadership Computing Facility, Argonne National Laboratory, Lemont, IL, USA
}
\date{}
\maketitle
\begin{abstract}
LLM agents are often evaluated by reducing execution traces to labels such as final-answer correctness, tool verification, and fault propagation. These labels are measurements produced by parsers, behavioral detectors, and harness rules, yet evaluations rarely report whether a measurement actually fits the trace it describes. We study what happens when those measurements do not fully capture the trace's behavior. Agents solve GSM8K problems with a calculator whose outputs are occasionally corrupted, which gives ground truth for what the tool returned, what the model later used, and what the detectors reported; a chemistry task with search and retrieval tests the case where the model cannot recompute the tool's output. Across eight models and up to 1,000 traces per model, final correctness, visible re-verification, and fault propagation disagree in systematic ways. The most common gap is the silent tool-output override: the model replaces a corrupted value without calling the tool again, so the error stops propagating while a verification detector records no re-check. Human review finds verification labels the least reliable ($\kappa=0.29$), and checking depends on the task: the same calculator that several models re-checked on arithmetic was never re-checked on chemistry. Execution limits and serving failures also change what is measured. These findings motivate measurement transparency: an automated label should be reported with its evidence, whether the detector's question applied, whether the run was censored, and any human corrections. Our trace viewer makes this information inspectable alongside the reported results.

\end{abstract}

\section{Introduction}
\label{introduction}
Agent evaluation is a measurement process. An agent produces an execution trace--a record of its model and tool interactions--which a parser or detector maps to a label. Labels are then aggregated into numbers such as accuracy, verification rate, or fault-propagation rate. Those numbers are often treated as facts about the agent. In practice, they also depend on the instruments used to measure the agent.

This distinction matters because agents are increasingly evaluated for tool use and multi-step decision-making. In such settings, a small measurement mistake can change the interpretation of a run. A trace may contain a correct answer that the parser does not extract. A detector may ask a question that is not meaningful for a particular event. A step limit or a serving failure may stop an agent before it finishes, but the benchmark may still count the run as simply wrong. In higher-stakes settings, such as clinical, defense, or other safety-critical workflows, this kind of false confidence is especially dangerous: an agent can look most trustworthy exactly when it is quietly overriding its instruments. Knowing that an agent received a score is not enough; we also need to know whether the score reflects a trustworthy measurement.

Recent work has already shown that the reliability of the evaluator cannot be assumed. Agent benchmark audits find disagreement between automated evaluators and human annotators, tool-calling benchmarks contain brittle judging rules, and agent performance can depend on the harness in which the model is run \citep{bean2025measuring,gurram2026evaluating,bhat2026benchmarking,yao2026harness}. We build on that line of work, but focus on a narrower question: when an automated trace detector reports a label, did the trace actually contain the right evidence for that label to be interpreted safely?

We study the question first in a deliberately simple domain. Agents solve arithmetic word problems using a calculator tool. The calculator sometimes returns a plausible but wrong value. Because arithmetic has external ground truth, we can check four things separately: what the calculator should have returned, what it actually returned, what value the model later used, and what final answer the model gave. We then test whether the patterns hold where the model cannot redo the work itself, using a chemistry task in which answers depend on retrieved molecular energies. 

The motivating failure is silent tool-output override. A model receives a corrupted calculator result, but later uses a different value, often the independently correct one, without calling the calculator again and without explicitly stating that the tool was wrong. Prior work on tool-memory conflict already shows that models arbitrate between internal knowledge and tool outputs \citep{cheng2025tool,varlamov2026memtoc}. Our claim is not that this behavior is newly discovered, but that it creates a measurement problem. For example, if the calculator returns 29 for a quantity whose correct value is 30 and the model later continues its reasoning with 30 without another tool call, a propagation detector may correctly record that 29 was not reused while a verification detector still reports no visible re-check. The trace therefore contains evidence of recovery that the verification detector was not designed to represent.

%%This paper argues for measurement transparency in agent evaluation. We should not only report labels; we should also report the state of the measurement that produced them. Was the detector applicable? What evidence did it use? Did a harness limit censor the run? Did a human reviewer correct the automatic label? These questions are especially important when labels are later used to compare models, choose a model-harness configuration, or produce feedback for improving future detectors.

\textbf{Contributions.}

\begin{enumerate}[label=\textbf{C\arabic*.},ref=C\arabic*, leftmargin=*]
\item \label{contrib:C1}
\textbf{Measurement transparency for trace detectors.}
We define a practical reporting principle for agent evaluation: a detector label should be accompanied by the evidence that supports it, whether the detector was applicable, whether execution was censored, and any known measurement risks or human corrections.

\item \label{contrib:C2}
\textbf{A controlled audit of detector reliability.}
In a pilot of 30 problems per model and a scaled audit of up to 1000 traces per model, we show that final correctness, visible verification, and fault propagation can diverge in systematic ways, including silent override, inconclusive propagation, formatting failures, parser loss, and censoring from execution limits and from the serving stack. A second task with three tools shows that verification behavior depends on whether the tool's output can be checked. We also compare automated labels with human review and report all estimates with uncertainty.

\item \label{contrib:c3}
\textbf{A trace viewer that reveals the measurement behind the score.}
We build a reliability-aware visualization that links aggregate results to the trace evidence behind each detector output and preserves automatic labels together with human corrections. This makes it possible to inspect why a measurement should or should not be trusted rather than treating the final score as self-explanatory.

\end{enumerate}

\section{Related Work}
\textbf{Evaluator validity.}
Construct-validity work argues that a benchmark score is meaningful only when the task and metric support the claim being made from them \citep{bean2025measuring}. AgentProp-Bench compares automated judging protocols with human labels and studies error propagation in tool-using agents
\citep{gurram2026evaluating}. \citet{bhat2026benchmarking} audit several
tool-calling benchmark families and report evaluator-human disagreements, brittle deterministic rules, judge variance, and corrected benchmark components. REFLECT similarly meta-evaluates LLM judges using controlled interventions on agent traces and finds substantial failure in fine-grained failure detection \citep{wang2026reflect}. AgentJudgeBench studies judge
reliability specifically for structured tool-calling workflows \citep{verma2026agentjudgebench}. Our work is complementary: rather than evaluating an LLM judge as a whole, we focus on behavioral trace detectors
and ask whether a detector had the observable and applicable evidence needed to support its label.

\textbf{Trace- and process-level evaluation.}
Recent work increasingly evaluates how an agent reaches an outcome rather than only whether it succeeds. ClawTrack separates task success from
process quality and includes result verification as a process dimension \citep{wu2026clawtrack}. CTF-ABACUS reconstructs agent trajectories to distinguish demonstrated exploitation from shortcut or unsupported success in cybersecurity tasks \citep{milner2026ctfabacus}. These works reinforce
the value of trace-level evidence. Our question is one level lower: whether the measurement applied to that trace faithfully represents the behavior
it is intended to measure.

%%\textbf{Tool-output conflict.} Tool-Memory Conflict and related work study cases where model knowledge and tool output disagree \citep{cheng2025tool,varlamov2026memtoc}. We do not rename that phenomenon or claim to discover it. Instead, we ask what happens when such conflict appears inside an evaluation pipeline. A model may recover from a bad tool value in a way that improves the final answer while still failing the detector's narrow definition of verification.

\textbf{Faulty tools and conflicting evidence.} \citet{sun2024toolsfail} show that strong models can reject faulty calculator outputs when the task is framed as detecting them; we measure how often they do so unprompted, during ordinary tool use. Tool-memory conflict studies cases where model knowledge and tool output disagree \citep{cheng2025tool,varlamov2026memtoc}, and work on knowledge conflicts finds that models adopt coherent external evidence over their own correct answer \citep{xie2024chameleon}, more readily the less confident they are \citep{wu2024clasheval}. We do not claim to discover these behaviors; we ask what they do to an evaluation pipeline, including when the model has no prior to weigh against the tool at all.

\textbf{Agent behavior under faulty tool outputs.} A growing set of studies injects wrong or misleading tool outputs and measures how agents respond. \citet{zhang2026} corrupt search and calculator feedback and find that persistent misleading feedback can leave an agent worse off than receiving no feedback at all. \citet{mittal2026} perturbs a single tool output in otherwise successful trajectories and finds that most runs recover, often without acknowledging the earlier error, and that fluent wrong evidence is rarely noticed. ToxicBench \citep{tao2026} rewrites successful tool observations for data agents and reports how often agents comply blindly, flag a problem, run a check, or adopt the corrupted value even after checking. \citet{kaynar2026} inject faults into a sandboxed multi-tool environment and find that agents often build confident answers on faulted data. These studies characterize what agents do when a tool is wrong. Our focus is complementary: we ask whether the automated labels used to report those behaviors, such as ``verified'' or ``recovered,'' match what the trace actually shows.

%%\textbf{Harness effects.} Harness-Bench shows that agent performance should often be reported at the model-harness configuration level rather than as a property of the model alone \citep{yao2026harness}. Here, the evaluation harness--the runtime configuration governing execution limits and tool behavior—matters because step and token limits can censor traces, while per-call fault injection gives different fault exposure to agents that make different numbers of tool calls. These are not just engineering details; they change what the reported denominator means.

\textbf{Harness effects.} Harness-Bench shows that agent performance should often be reported at the model-harness level rather than as a property of the model alone \citep{yao2026harness}. In our setting, step and token limits can censor traces, and per-call fault injection gives different fault exposure to agents that make different numbers of tool calls; both change what a reported denominator means.

\textbf{Trace visualization.}
Systems such as AgentLens and Graph of Trace make agent behavior easier to inspect \citep{lu2024agentlens,gao2026graph}. Agentic CLEAR performs system-, trace-, and node-level agent evaluation and presents the resulting analysis through an interactive interface \citep{yehudai2026agenticclear}.
Harness Lab likewise supports trace-level inspection and human adjudication for benchmark audits \citep{bhat2026benchmarking}. Our viewer is not
presented as the first trace-analysis interface. Its focus is the measurement layer: detector outputs are shown together with their evidence, applicability, censoring state, risk flags, and preserved human corrections.

\section{Experimental Setup}
\subsection{Task and fault injection}
\label{sec:tasks}
\textbf{Arithmetic.} Agents solve GSM8K \citep{cobbe2021gsm8k} problems in a ReAct-style loop with a calculator: the model reasons, may call the tool, observes the result, and continues until it writes \texttt{FINAL ANSWER:} followed by a number. The prompt instructs the model to use the calculator for arithmetic; it is not told that the tool may be wrong. Each calculator call is corrupted with probability 0.15. An integer result is replaced by one of the true value $\pm1$, $\pm2$, or an adjacent-digit transposition; a non-integer result is shifted by 5\%. Faults are drawn from a per-question random stream seeded with $\mathrm{sha256}(\textit{seed}{:}\textit{index})$, so a question receives the same fault stream under a given seed regardless of execution order or concurrency. The \emph{pilot} audit uses the first 30 problems of the GSM8K training split with seed 50; the \emph{scaled} audit uses the first 500 problems under seeds 50 and 51 (1,000 traces per model), so the pilot problems and seed are subsets of it. All runs use temperature 0 (except the scaled claude-sonnet-5 run, \S3.2) and 1,024 output tokens per turn.

\textbf{Chemistry.} To test a setting in which the model cannot recompute a tool's output, the agent has three tools: \texttt{search\_molecule} (name $\rightarrow$ SMILES), \texttt{retrieve\_property} (SMILES, property, method, temperature $\rightarrow$ value), and the same calculator. Values come from a knowledge base of 57 records for 22 molecules built from the public logs \citep{chemgraphlogs} of the ChemGraph agent \citep{pham2025chemgraph}: energies and dipole magnitudes computed with MACE-MP \citep{batatia2023mace} and GFN2-xTB \citep{bannwarth2019gfn2}. The 280 questions are the 27 original ChemGraph questions and 253 generated ones (lookups, differences between molecules or methods, Gibbs-energy changes between temperatures, and reaction energies needing three to six lookups); gold answers are computed from the same records the tool serves (details in Appendix~\ref{app:chem}). Each search, retrieval, and calculator call is faulted with probability 0.15: search returns a different molecule's SMILES, retrieval returns the value for a neighboring condition or molecule or a value perturbed by 0.5--3\%, and the calculator behaves as above. Each model runs seeds 50 and 51 with faults and seed 50 without (a clean control).

\textbf{Instrument check.} Every trace records the tool inputs, the returned and the true outputs, and the fault flag. Replaying every recorded tool call from its seed reproduces all of them in the scaled audit (10,881 traces that reached a tool call), so the injector does what the analysis assumes.

\subsection{Models}
The pilot covers six models: Mistral-Large-3-675B, nemotron-3-ultra, claude-sonnet-5, claude-opus-4.8, gpt-oss-120b, and gpt-oss-20b. The scaled arithmetic audit adds Llama-3.1-70B and Gemma-4-31B and includes both Claude models. The gateway no longer accepts a temperature setting for claude-sonnet-5, so its scaled run uses the provider's default sampling; every other run uses temperature 0. The chemistry task covers gpt-oss-120b, gpt-oss-20b, Gemma-4-31B, Llama-3.1-70B, and Mistral-Large-3-675B. Open models are served through OpenAI-compatible vLLM endpoints on a shared institutional inference service, and the Claude models through an institutional API gateway. All scaled runs use one agent loop with \texttt{MAX\_STEPS}=50 and \texttt{MAX\_TOKENS}=1024; the pilot Claude runs used a separate Anthropic-API loop with a 15-step limit, which no Claude trace approached (at most 6 steps).

\subsection{Detectors under audit}

Table~\ref{tab:detectors} gives the decision rule of each automatic measurement. The detectors are instruments, not ground truth.

\begin{table}[h]
\centering
\caption{Automatic measurements audited in this work and their decision rules.}
\label{tab:detectors}
\begin{tabularx}{\textwidth}{@{}l Y Y@{}}
\toprule
\textbf{Detector} & \textbf{Decision rule} & \textbf{Why the label may need qualification} \\
\midrule
D-correct &
Parse the number after the last \texttt{FINAL ANSWER:}; compare with gold (exact for GSM8K, a stated tolerance for chemistry). Runs stopped by a limit are \emph{censored}; runs ended by a serving error are \emph{errors} and excluded from denominators. &
A correct answer is missed if the format defeats the parser (F2). \\[3pt]
D-verify &
\emph{Re-check}: a step calls an expression (whitespace-normalized) already called in an earlier step. \emph{Claimed only}: no such repeat, but the step's text contains a verification phrase (``verify'', ``double-check'', ``make sure''). &
Recovery without a second identical call is invisible; repeated calls in a loop look like verification. \\[3pt]
D-impact &
For each faulted call, scan the next 30 steps for the faulted and the true value: as a number in a later tool input, through a later expression that contains the faulted sub-expression, in later text, or as the final answer. Faulted value reaches the answer, or appears without the true value: \emph{used}. True value appears: \emph{replaced}. Neither: \emph{inconclusive}. &
If neither value reappears, propagation is undecidable (F3). \\[3pt]
Fault relevance &
A fault is relevant if any D-impact signal links either value to later computation or the final answer. &
Derived from D-impact; inherits its blind spots. \\[3pt]
Quiet replacement &
A \emph{replaced} event with no later re-run of the same call and none of 19 noticing phrases in the next two thoughts. &
Keyword-based; cannot see reasoning the model does not return. \\
\bottomrule
\end{tabularx}
\end{table}

%%\subsection{System Workflow}

%%Figure~\ref{fig:system-workflow} summarizes where the measurement layer fits within the evaluation pipeline. Agent-tool interactions first produce structured traces, which are processed by the three automated detectors. Their outputs are then exposed through the terminal summary and interactive viewer for inspection and human review.
%%\begin{figure}[t]
%%\centering
%%\includegraphics[width=0.50\linewidth]{Trace-System overview.jpg}
%%\caption{\textbf{System workflow.}
%%GSM8K questions and seeded calculator faults produce agent execution traces.
%%The measurement layer applies D-correct, D-verify, and D-impact to generate
%%automatic labels and risk flags. The review layer exposes these outputsthrough a terminal summary and interactive viewer for trace inspection andhuman comparison.}
%%\label{fig:system-workflow}
%%\end{figure}

\subsection{Measurement states}
For each detector output, we store the automatic label, supporting evidence, applicability, censoring state, risk flags, and any human correction. Human corrections are kept alongside rather than replacing the original output (\textbf{\ref{contrib:C1}}).

Table~\ref{tab:measurement} groups the observed issues into seven practical failure classes, each with where it appears in this paper. The taxonomy is empirical rather than exhaustive.

%%\begin{table}[h]
%%\centering
%%\caption{Observed measurement issues in the study.}
%%\label{tab:measurement}
%%\begin{tabularx}{\textwidth}{@{}l l Y@{}}
%%\toprule
%%\textbf{Class} & \textbf{Issue} & \textbf{Description} \\
%%\midrule
%%F1 & Detector blind spot &
%%The behavior occurs in the trace, but the detector's rule cannot represent it. \\[4pt]
%%F2 & Parser loss &
%%The model gives a usable answer, but extraction fails. Example: ''192 million'' required a unit-aware rule.\\[4pt]
%%F3 & Inapplicable question &
%%The detector asks a question that the trace does not answer. \\[4pt]
%%F4 & Formatting sensitivity &
%%Small formatting differences change the detector verdict. Example: \texttt{184-145} and \texttt{184 -145} can yield different detector behavior. \\[4pt]
%%F5 & Instrument assumption &
%%The fault injector or measurement tool was assumed correct before being checked. \\[4pt]
%%F6 & Rate uncertainty &
%%A single batch percentage is treated as a stable rate without uncertainty. \\[4pt]
%F7 & Censoring &
%%A run is cut off by a step or token limit or serving failure and then %%scored like an ordinary failure. \\
%%\bottomrule
%%\end{tabularx}
%%\end{table}

\begin{table}[t]
\centering
\caption{Observed measurement issues and where each is evidenced.}
\label{tab:measurement}
\small
\begin{tabularx}{\textwidth}{@{}l l Y@{}}
\toprule
\textbf{Class} & \textbf{Issue} & \textbf{Description and evidence} \\
\midrule
F1 & Detector blind spot & Behavior the rule cannot represent: quiet replacement is 10--85\% of decidable fault events (\S\ref{sec:scale}). \\
F2 & Parser loss & A usable answer is not extracted: Llama omits the answer line on 25 of 840 chemistry traces, 17 of which state the correct value in prose; scaled answers (``192 million'') required a unit-aware rule. \\
F3 & Inapplicable question & The trace cannot answer the detector's question: inconclusive D-impact on 12 of 17 (gpt-oss-20b) and 56 of 155 (nemotron) pilot fault events (\S\ref{sec:pilot}). \\
F4 & Formatting sensitivity & \texttt{200-142-16} tokenizes as $[200,-142,-16]$, hiding a reused faulted 142 (one reviewed trace, \S\ref{sec:human}). \\
F5 & Instrument assumption & A component trusted without a check: the fault injector (verified by replay, \S\ref{sec:tasks}); serving-side tool-call parsers that return calls as text (\S\ref{sec:chem}). \\
F6 & Rate uncertainty & A single run's rate treated as stable: identical reruns at temperature 0 flip correctness for 8\% of one model's questions (\S\ref{sec:scale}). \\
F7 & Censoring & A run cut off by a step limit or serving failure and scored like an ordinary failure (\S\ref{sec:scale}, \S\ref{sec:ablation}). \\
\bottomrule
\end{tabularx}
\end{table}

\subsection{Measurement-transparent inspection}
\label{visualization}

The inspection workflow for \textbf{\ref{contrib:c3}} has two levels. A terminal summary gives a compact overview of outcomes, fault exposure, verification events, and flagged cases for triage (Appendix~\ref{app:viewer}). The interactive viewer then aligns traces by question and model and exposes, for any trace, the tool outputs, the subsequent model behavior, the automatic labels, and any human correction (Figure~\ref{fig:viewer-trace}; a larger corpus view is in Appendix~\ref{app:viewer}).
 
\begin{figure}[t]
\centering
\includegraphics[width=\linewidth]{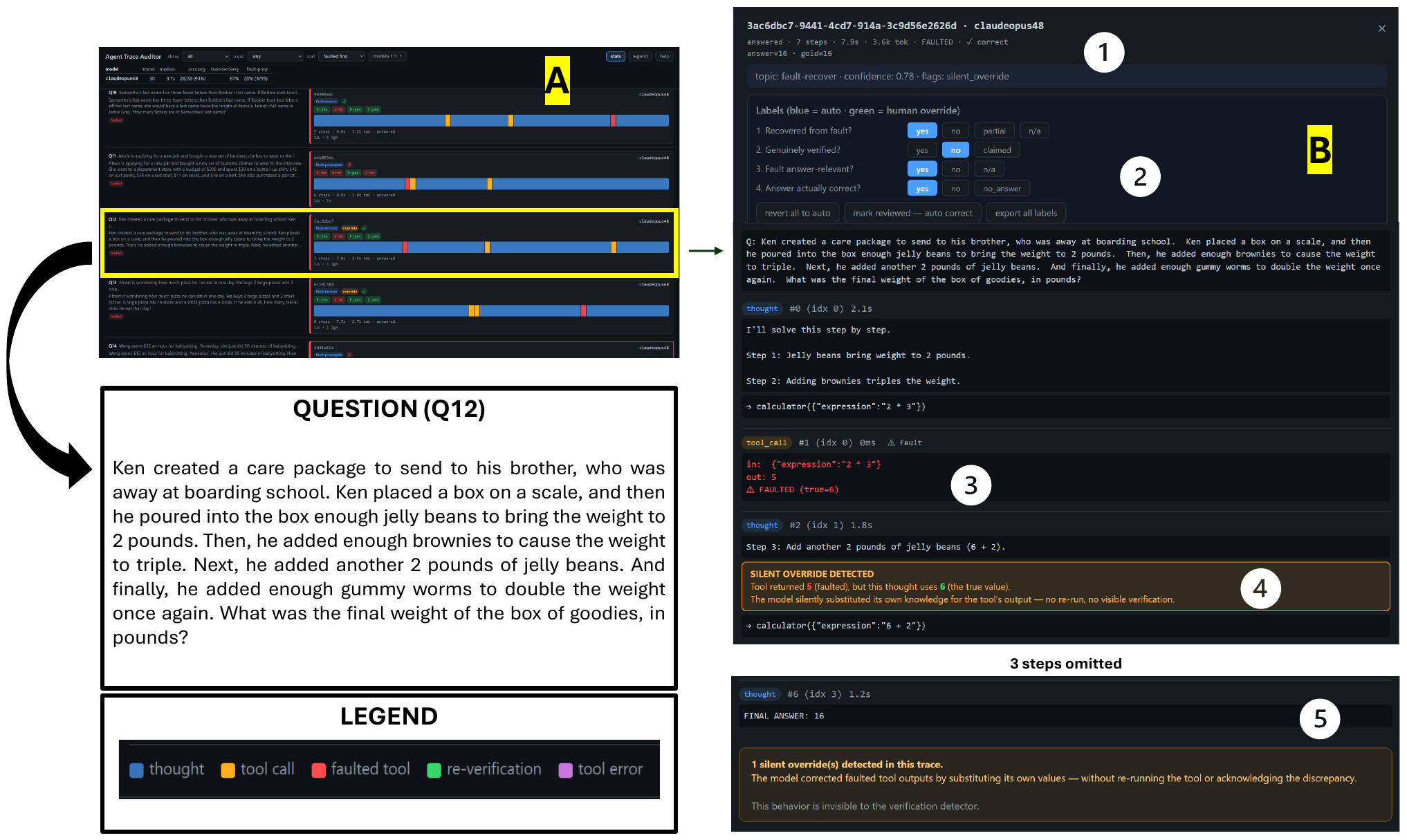}
\caption{\textbf{The trace viewer on a silent tool-output override.} \textbf{(A)} Corpus view of claude-opus-4.8's pilot run (30 traces), with per-model accuracy, fault-recovery and fault-propagation rates in the header. Each card shows the automatic labels as chips (R recovered, V verified, F fault answer-relevant, C correct; green yes, red no) and a timeline of steps colored by the legend, with block width proportional to duration. The highlighted card is expanded in (B), and its question is reproduced below the panel. \textbf{(B)} Trace view of the same trace, with three intermediate tool steps omitted. \textcircled{\scriptsize 1} The trace was faulted, yet its final answer matches the gold answer. \textcircled{\scriptsize 2} The automatic labels record recovery but not verification; a reviewer can override any label (shown in green, with the automatic value kept and restorable) and export the corrected labels. \textcircled{\scriptsize 3} The calculator returns a corrupted 5 for $2\times 3$ (true value 6). \textcircled{\scriptsize 4} The next thought uses 6 without re-running the tool or noting the discrepancy, and the viewer flags a silent override. \textcircled{\scriptsize 5} The answer is correct, so an outcome-only check scores the trace as a success, while the verification detector records no re-check.}
\label{fig:viewer-trace}
\end{figure}

%%\begin{figure}[t]
%%\centering

%%\begin{subfigure}{0.95\linewidth}
   %% \centering
  %%  \includegraphics[width=\linewidth]{Trace_interface.png}
  %%  \caption{Corpus-level comparison across questions and models.}
  %%  \label{fig:viewer-corpus}
%%\end{subfigure}

%%\vspace{2mm}

%%\begin{subfigure}{0.80\linewidth}
    %%\centering
   %% \includegraphics[width=\linewidth]{silent overide example.png}
   %% \caption{Trace-level inspection of a silent tool-output override.}
   %% \label{fig:viewer-trace}
%%\end{subfigure}

%%\caption{\textbf{Interactive trace inspection.}
%%he corpus view supports comparison across model runs, while the trace view exposes the evidence behind an individual detector output. In panel (b), the tool returns a corrupted value that the model later replaces without an explicit tool re-check.}
%%\label{fig:viewer}
%%\end{figure}

\section{Results}

%%This section presents the controlled detector audit in \textbf{\ref{contrib:C2}}. We first compare detector outputs across six model families as shown in Table~\ref{tab:cross-model}, then examine trace patterns that make those outputs difficult to interpret, test the effect of harness constraints, and finally compare the automated labels with human review. 

We first report the 30-question pilot (\textbf{\ref{contrib:C2}}), then the scaled audit, the chemistry task, the harness ablation, and the comparison with human review.

\subsection{Pilot audit}
\label{sec:pilot}
Table~\ref{tab:cross-model} summarizes the pilot (seed 50, first 30 GSM8K training problems). Verification counts traces with a tool re-check, with traces that only claimed to check in parentheses; quiet replacement counts traces with at least one such event. These counts come from single runs on 30 problems, so differences of a few traces are within run-to-run variation (\S\ref{sec:scale}); the scaled audit gives the model-level rates.

\begin{table}[!ht]
\centering
\caption{Pilot audit: 30 problems per model, seed 50, separate runs from Table~\ref{tab:ablation}. Outcome is completed-correct/completed-wrong/not-completed (step-limit or no final answer); serving errors are excluded, as in all other tables (Mistral: 8). Verify and Quiet repl.\ count traces: a trace counts if it has at least one quiet replacement under the revised D-impact and the rule of Table~\ref{tab:audit}, which counts fault events instead. Outcomes now apply one rule to every model, which moves one gpt-oss-20b and one Mistral trace relative to the first version of this paper.}
\label{tab:cross-model}
\small
\begin{tabularx}{\textwidth}{@{}l l l l Y@{}}
\toprule
\textbf{Model} & \textbf{Outcome} & \textbf{Verify} & \textbf{Quiet repl.} & \textbf{Measurement caveat} \\
\midrule
claude-sonnet-5 & 30 / 0 / 0 & 5/30 & 6/30 & No censoring or errors; recoveries visible only at the trace level (F1). \\
claude-opus-4.8 & 28 / 2 / 0 & 2/30 (+3) & 9/30 & Three traces only claim to verify (D-verify's claimed-only label). \\
gpt-oss-120b & 28 / 2 / 0 & 8/30 & 1/30 & Only 8 decidable fault events: one trace moves the rate by several points (F6). \\
gpt-oss-20b & 24 / 3 / 3 & 4/30 & 1/30 & 12 of 17 fault events inconclusive under the pilot detector (F3). \\
Mistral-Large-3-675B & 20 / 1 / 1 & 0/22 (+1) & 5/22 & Eight serving errors; recoveries with no visible re-execution (F1, F7). \\
nemotron-3-ultra & 20 / 0 / 10 & 11/30 & 20/30 & Every miss is a step-limit hit (F7); 56 of 155 fault events inconclusive (F3). \\
\bottomrule
\end{tabularx}
\end{table}

\textbf{Silent override and inconclusive propagation}. The clearest gap is silent override (Figure~\ref{fig:viewer-trace}): the calculator returns a corrupted value, and the trace later uses a different, often correct, value without a re-check. D-impact records non-propagation while D-verify records no verification. The detector is working as defined while asking a narrower question than the behavior of interest: visible re-execution is not the same as recovery. The opposite problem occurs when neither value reappears in a detectable form, and D-impact cannot decide propagation. Both are valid detector outputs that are incomplete descriptions of the trace.

%%The clearest measurement gap is silent tool-output override. A calculator returns a corrupted value, but the trace later uses a different value, often the independently correct one, without a calculator re-check. D-impact may therefore record non-propagation while D-verify records no visible verification. Figure~\ref{fig:viewer-trace} shows a representative example: the calculator returns a corrupted value, but the model later continues with the correct value without rerunning the tool. The recovery is visible in the trace even though D-verify records no explicit re-check. The detector can therefore be working as defined while still asking a narrower question than the behavior of interest: visible re-execution is not the same as recovery.

%%The opposite problem occurs when neither the corrupted nor corrected value reappears in a detectable form. In those cases, D-impact lacks enough evidence to determine propagation. Both cases illustrate \textbf{\ref{contrib:C2}}: a technically valid detector output can still be an incomplete description of the trace.

\subsection{Scaled audit}
\label{sec:scale}
Table~\ref{tab:audit} reports the scaled arithmetic audit with 95\% intervals. Intervals in the table are Wilson intervals over traces; Appendix~\ref{app:stats} gives question-clustered bootstrap intervals, which account for the two seeds sharing questions, and per-trace fault outcomes with and without looping traces.
 
\begin{table}[t]
\centering
\caption{Scaled arithmetic audit (500 problems $\times$ 2 seeds per model). Coverage excludes serving errors. Censored: runs stopped by the step limit or without an answer. First call: accuracy when the trace's first calculator call is clean vs.\ faulted (95\% Wilson intervals). Used: decidable fault events in which the faulted value was used. Quiet: fault events replaced with no re-run and no verbal flag (Table~\ref{tab:cross-model} counts the same behavior per trace). Re-checked: traces with a tool re-check.}
\label{tab:audit}
\small
\resizebox{\linewidth}{!}{%
\begin{tabular}{@{}lrrrcrrr@{}}
\toprule
Model & Coverage & Censored & Accuracy & First call: clean / faulted & Used & Quiet & Re-checked \\
\midrule
claude-opus-4.8 & 100\% & 0 & 93.7 [92.0, 95.0] & 94.6 [93, 96] / 88.0 [82, 92] & 43/460 & 260 & 62 \\
claude-sonnet-5$^\ast$ & 100\% & 0 & 92.9 [91.1, 94.3] & 93.9 [92, 95] / 86.6 [80, 91] & 42/385 & 138 & 128 \\
Gemma-4-31B & 100\% & 2 & 94.5 [92.9, 95.8] & 95.2 [94, 96] / 90.1 [84, 94] & 16/602 & 254 & 160 \\
gpt-oss-120b & 100\% & 5 & 90.2 [88.2, 91.9] & 91.6 [90, 93] / 82.9 [76, 88] & 31/387 & 87 & 182 \\
gpt-oss-20b & 100\% & 14 & 87.4 [85.2, 89.3] & 90.7 [89, 92] / 81.8 [74, 87] & 14/317 & 58 & 148 \\
Mistral-Large-3-675B & 66\% & 0 & 87.0 [84.2, 89.3] & 88.8 [86, 91] / 75.8 [66, 83] & 23/237 & 201 & 6 \\
nemotron-3-ultra & 100\% & 79 & 83.1 [80.7, 85.3] & 87.7 [85, 90] / 79.3 [72, 85] & 23/1006 & 101 & 278 \\
Llama-3.1-70B & 100\% & 0 & 69.9 [67.0, 72.7] & 77.4 [75, 80] / 23.6 [17, 31] & 100/159 & 59 & 0 \\
\bottomrule
\end{tabular}}
 
\vspace{2pt}
{\footnotesize $^\ast$Provider-default sampling: the gateway no longer accepts a temperature setting for this model.}
\end{table}

\textbf{Measuring the fault effect without exposure bias.} Comparing traces that received a fault with traces that did not is misleading: each call is faulted independently, so traces that make more calls, which are the ones that loop or struggle, are more likely to be faulted. We instead condition on whether the \emph{first} calculator call was faulted, a random draw made before the trace can diverge. For nemotron-3-ultra, which loops most, the naive split gives a gap of roughly 26 points and the first-call comparison 8.4 points: a threefold overstatement produced by the measurement, not the model.
 
\textbf{Models differ in how much they trust the tool.} The first-call effect is positive for every model and ranges from 5.2 points (Gemma) to 53.9 (Llama; 77.4\% vs.\ 23.6\%). Its question-clustered bootstrap interval excludes zero for every model except Gemma, $[-0.1, 11.0]$ (Appendix~\ref{app:stats}). Apart from Llama, the intervals overlap, so we do not rank the other models. Llama never re-ran a calculation and used the faulted value in 63\% of decidable events (100/159), while Gemma re-ran a calculation in 16\% of traces and used the faulted value in 2.7\% (16/602). Aggregate accuracy on clean problems separates neither model from the rest; only the per-fault view does.
 
\textbf{Silent override persists at scale.} Quiet replacements occur for every model, from 10\% of decidable fault events for nemotron-3-ultra (101/1,006) to 85\% for Mistral-Large-3-675B (201/237), and 57\% for claude-opus-4.8 (260/460). For Mistral almost every recovery is one D-verify cannot see. Because the verbal flag is a keyword heuristic over returned text, these shares are upper bounds on truly silent replacement for models whose reasoning is not returned.
 
\textbf{The detector version changes the reported result.} After the pilot's first human-review pass, D-impact was extended to recognize recovery through compound expressions and through values restated in the model's text (\S\ref{sec:human}). Re-scoring the \emph{same} pilot traces with the revised detector reduces inconclusive fault events from 12 of 17 to 2 of 17 for gpt-oss-20b and from 56 of 155 to 6 of 155 for nemotron-3-ultra, with accuracy and fault counts unchanged; at scale, 1--11\% of events remain inconclusive. Table~\ref{tab:cross-model} is kept as a record of the pilot detector; all scaled results use the revised one.
 
\textbf{Temperature 0 is not reproducible on shared endpoints.} During the scaled runs, 948 questions were executed twice under identical conditions for two models. The runs agreed on correctness for 511 of 516 questions (99.0\%) for Mistral but 398 of 432 (92.1\%) for gpt-oss-20b. Per-trace labels from one run are not stable facts about a model--question pair (F6), which also explains why the separately run baseline cells of Tables~\ref{tab:cross-model} and~\ref{tab:ablation} differ by a trace or two. We kept one run per question by an outcome-independent rule (Appendix~\ref{app:repro}).
 
\textbf{Serving-side censoring.} For Mistral, 34\% of traces (340/1,000) ended in a server-side error. The harness requires a tool call on the first turn; in 322 of the 340 errors the model replied in prose instead, and 302 were reported by the endpoint as tool-call parse failures. This is a harness--endpoint interaction rather than a model failure; 94\% of these errors (319/340) occurred before any tool output, so they cannot depend on the fault condition. Mistral's estimates are therefore computed on the 66\% of traces that completed. Execution limits matter too: nemotron-3-ultra is correct on 83.1\% of traces but 95.3\% of those that produced an answer, because 79 runs hit the step limit and 49 stopped without an answer (F7).
 
\subsection{A second task: chemistry with search and retrieval}
\label{sec:chem}
Table~\ref{tab:chem} summarizes the chemistry task. Accuracy falls by 27--33 points between the clean control and the faulted runs for the four models with full coverage, and by far more when the first retrieval is faulted.
 
\begin{table}[t]
\centering
\caption{Chemistry task (280 questions; faulted: 2 seeds, clean: 1 seed). First call: accuracy when the first retrieval is clean vs.\ faulted. Used: decidable fault events in which the bad value was used, per tool.}
\label{tab:chem}
\small
\begin{tabular}{@{}lccccc@{}}
\toprule
 & \multicolumn{2}{c}{Accuracy (\%)} & First call (\%) & \multicolumn{2}{c}{Used the bad value} \\
\cmidrule(lr){2-3}\cmidrule(lr){5-6}
Model & Clean & Faulted & clean $\rightarrow$ faulted & Retrieval & Calculator \\
\midrule
gpt-oss-120b  & 99.6 & 66.4 & 79.2 $\rightarrow$ 3.2 & 155/156 & 62/62 \\
Gemma-4-31B   & 100.0 & 68.6 & 81.9 $\rightarrow$ 4.3 & 156/159 & 42/58 \\
Llama-3.1-70B & 93.2 & 65.9 & 75.2 $\rightarrow$ 4.1 & 143/147 & 40/53 \\
Mistral-Large-3-675B & 93.6 & 65.5 & 79.6 $\rightarrow$ 2.9 & 159/172 & 14/60 \\
gpt-oss-20b$^\dagger$ & 97.3 & 73.3 & 85.7 $\rightarrow$ 2.2 & 68/68 & 6/27 \\
\midrule
All & & & & 681/702 (97\%) & 164/260 (63\%) \\
\bottomrule
\end{tabular}
 
\vspace{2pt}
{\footnotesize $^\dagger$Over traces the endpoint served correctly (coverage 53\% clean, 59\% faulted; see text).}
\end{table}
 
\textbf{Retrieved values are trusted almost without exception.} A wrong retrieved value was used in 681 of 702 decidable cases, and a faulted lookup was re-run in only 2 of 718 fault events. When the first retrieval was faulted, accuracy fell to 2--4\% for every model. A retrieved energy offers nothing to check against, so the absence of verification is a structural limit rather than a lapse: the error is visible only in the trace.
 
\textbf{The same calculator is trusted far more.} Table~\ref{tab:calc-domains} compares the unchanged calculator across tasks. On arithmetic, three models re-ran a faulted calculation in 41--70\% of fault events; on chemistry, no model re-ran one, and the share of decidable cases in which a wrong result was used rose from 10.8\% (184/1,702) to 63.1\% (164/260). The tasks differ in operands (small integers vs.\ energies such as $-46.2465$\,eV) and in the fault applied ($\pm1$, $\pm2$, or a transposition vs.\ a 5\% shift), so we claim the direction rather than the size of the effect: the re-checking several models did routinely on arithmetic disappears once the numbers cannot be recomputed in the model's head. Verification measured on easy tasks is not evidence of verification where it matters.
 
\begin{table}[t]
\centering
\caption{The calculator across tasks. Used: wrong result used, over decidable events. Re-ran: faulted call re-run, over all calculator fault events.}
\label{tab:calc-domains}
\small
\begin{tabular}{@{}lcccc@{}}
\toprule
 & \multicolumn{2}{c}{Arithmetic} & \multicolumn{2}{c}{Chemistry} \\
\cmidrule(lr){2-3}\cmidrule(lr){4-5}
Model & Used & Re-ran & Used & Re-ran \\
\midrule
gpt-oss-120b    & 31/387 (8\%)  & 270/413 & 62/62 (100\%)  & 0/62 \\
Gemma-4-31B     & 16/602 (3\%)  & 254/625 & 42/58 (72\%) & 0/58 \\
Llama-3.1-70B   & 100/159 (63\%) & 0/179  & 40/53 (75\%) & 0/64 \\
Mistral-Large-3-675B & 23/237 (10\%)  & 2/261   & 14/60 (23\%) & 0/67 \\
gpt-oss-20b     & 14/317 (4\%)  & 248/357 & 6/27 (22\%)  & 0/27 \\
\bottomrule
\end{tabular}
\end{table}
 
\textbf{Search faults are mostly bypassed, not detected.} A wrong molecule from search was used in only 10--21\% of decidable cases for the four models that called search regularly. We do not read this as detection: models often passed the molecule's name directly to retrieval, avoiding the fault without noticing it. Separating bypass from detection requires labels.
 
\textbf{Failures that are not the model's.} On gpt-oss-20b, 359 traces ended with the model's tool call returned as plain text (\texttt{to=functions.retrieve\_property \ldots <|call|>}) instead of a structured call, when the model requested several lookups in one turn and the endpoint's parser failed. An outcome-only evaluation scores gpt-oss-20b at 52\% on the clean control; on the traces served correctly, it scores 97\%. We count these as serving errors and report coverage (F5, F7).

\subsection{Harness constraint ablation}
\label{sec:ablation}

%%Harness limits provide a second test of measurement validity: a trace stopped before a final answer is censored, not equivalent to a completed but incorrect run. We vary \texttt{MAX\_STEPS} $\in \{25,50,150\}$ and \texttt{MAX\_TOKENS} $\in \{1024,4096\}$ for nemotron-3-ultra and gpt-oss-120b on the same 30 questions with fault seed 50.

%%Table~\ref{tab:ablation} shows different sensitivity across the two models. gpt-oss-120b remains between 27/30 and 29/30 correct, while nemotron-3-ultra ranges from 21/30 to 27/30. Candidate override is common for nemotron (15--24 traces per cell) but rare for gpt-oss-120b (0--3), showing that similar aggregate outcomes can hide different recovery behavior.

We vary \texttt{MAX\_STEPS} $\in \{25,50,150\}$ and \texttt{MAX\_TOKENS} $\in \{1024,4096\}$ for nemotron-3-ultra and gpt-oss-120b on the pilot problems, under seed 50 (the original run) and seeds 51 and 52 (Table~\ref{tab:ablation}). We compare each cell with the baseline (50 steps, 1,024 tokens) by exact McNemar tests paired by question and seed. gpt-oss-120b stays within 27--29 of 30 in every cell and seed, and no cell differs from the baseline. nemotron-3-ultra ranges from 21 to 28, but its lowest run is the seed-50 baseline itself (21/30), and both reruns of that cell score 27. Only one of its five comparisons with the baseline reaches $p<0.05$ (25 steps, 4,096 tokens: $p=0.039$), which does not survive a Bonferroni correction, and none of the token comparisons at a fixed step limit does (lowest $p=0.065$). What the budget changes is censoring: 59 of nemotron-3-ultra's 71 misses across the 18 runs are runs stopped by a limit, against at most one wrong answer per run. A single-seed ablation would have reported a budget effect that three seeds attribute mostly to run-to-run variation (F6). The nemotron-3-ultra reruns were served from a different cluster after the model moved, which may contribute to their lower censoring. Appendix~\ref{app:ablation} gives the seed-50 fault, verification and quiet-replacement counts.

\begin{table}[t]
\centering
\caption{Constraint ablation over three seeds (pilot problems, 30 per run). Correct and Not compl.\ (runs stopped by the step limit or without an answer) are given for seeds 50/51/52; serving errors are excluded. $p$: exact McNemar test against the baseline cell (steps=50, tokens=1024), paired by question and seed.}
\label{tab:ablation}
\small
\begin{tabular}{@{}l l c r c r@{}}
\toprule
\textbf{Model} & \textbf{Cell} & \textbf{Correct} & \textbf{Mean} & \textbf{Not compl.} & \textbf{$p$} \\
\midrule
\multirow{6}{*}{nemotron-3-ultra}
 & steps=25,  tokens=1024 & 23 / 26 / 26 & 25.0 & 7 / 3 / 3 & 1.000 \\
 & steps=25,  tokens=4096 & 27 / 28 / 27 & 27.3 & 3 / 1 / 2 & 0.039 \\
 & steps=50,  tokens=1024 & 21 / 27 / 27 & 25.0 & 9 / 2 / 2 & -- \\
 & steps=50,  tokens=4096 & 27 / 25 / 26 & 26.0 & 3 / 4 / 3 & 0.581 \\
 & steps=150, tokens=1024 & 24 / 28 / 27 & 26.3 & 6 / 1 / 2 & 0.289 \\
 & steps=150, tokens=4096 & 25 / 28 / 27 & 26.7 & 5 / 1 / 2 & 0.180 \\
\midrule
\multirow{6}{*}{gpt-oss-120b}
 & steps=25,  tokens=1024 & 27 / 29 / 28\textsuperscript{$\ddagger$} & 28.0 & 1 / 0 / 0 & 1.000 \\
 & steps=25,  tokens=4096 & 29 / 28 / 28 & 28.3 & 0 / 0 / 0 & 1.000 \\
 & steps=50,  tokens=1024 & 27 / 29 / 28 & 28.0 & 0 / 0 / 0 & -- \\
 & steps=50,  tokens=4096 & 28 / 29 / 29 & 28.7 & 0 / 0 / 0 & 0.500 \\
 & steps=150, tokens=1024 & 28 / 28 / 28 & 28.0 & 0 / 0 / 0 & 1.000 \\
 & steps=150, tokens=4096 & 27\textsuperscript{$\ddagger$} / 28 / 27 & 27.3 & 0 / 0 / 0 & 1.000 \\
\bottomrule
\end{tabular}
 
\vspace{2pt}
{\footnotesize $^\ddagger$Out of 29: one serving-error trace excluded, as in all scaled results.}
\end{table}

\subsection{Human adjudication and detector reliability}
\label{sec:human}
%%We hand-reviewed 90 of 180 traces across the six model families on four dimensions: fault recovery, verification behavior, fault relevance, and final-answer correctness. Human labels were compared with automated labels where the measurement was applicable, and corrections were stored without replacing the original detector output. Agreement therefore describes the reviewed comparable cases rather than all 180 traces.

%%Manual review also revealed coverage gaps in the automated detection logic, including traces with no obvious anomaly and cases where the initial rules did not capture relevant behavior. We revised the detector logic to cover these cases, reran the automated labels, and re-checked previously reviewed traces against the revised outputs. The agreement results below therefore compare the final detector implementation with the final set of human-reviewed labels.

One author reviewed 90 of the 180 pilot traces on four questions: did the trace recover from a fault, did it genuinely verify, was the fault relevant to the answer, and was the final answer correct. Traces were selected by the reviewer rather than by a sampling procedure: selection was guided by detector flags and extended to cover each behavior cluster, including clean traces with no fault. Agreement therefore describes the reviewed traces, not the full corpus, and likely over-represents traces the detectors found difficult. Human labels were stored without replacing the automatic ones. Automatic labels were visible during review, and for 67 of the 90 traces the reviewer confirmed all automatic labels in one step rather than answering each question separately, which favors agreement; the D-correct comparison comes entirely from these confirmations. Against the detectors as they were before the revision, the same labels give D-impact 14/14, fault relevance 14/18 (four relevant faults missed), and unchanged D-verify. The first review pass exposed gaps in D-impact (recovery through compound expressions, values restated in text, and the tokenization issue F4), so we revised D-impact and re-checked the reviewed traces; D-verify was not changed. Agreement is therefore measured against the revised D-impact on traces that informed its revision, and should be read as an upper bound for D-impact and as a direct measurement for D-verify.

\begin{table}[t]
\centering
\caption{Agreement between automatic labels and human review on reviewed pilot traces where the label is comparable (exact 95\% intervals). ``--'' marks $\kappa$ where all labels share one value.}
\label{tab:agreement}
\small
\begin{tabular}{@{}lccc@{}}
\toprule
\textbf{Detector} & \textbf{Agreement} & \textbf{Rate (95\% CI)} & \textbf{Cohen's $\kappa$} \\
\midrule
D-correct           & 67 / 67 & 100\% [94.6, 100] & 1.00 \\
D-impact / recovery & 19 / 19 & 100\% [82.4, 100] & 1.00 \\
Fault relevance     & 18 / 18 & 100\% [81.5, 100] & -- \\
D-verify            & 67 / 81 & 82.7\% [72.7, 90.2] & 0.29 \\
\bottomrule
\end{tabular}
\end{table}

D-verify is the only detector with observed disagreement (Table~\ref{tab:agreement}). Its raw agreement overstates its reliability, because most reviewed traces contain no verification (75 of 81 human labels are negative): $\kappa=0.29$ is only fair agreement beyond chance. Of the 14 disagreements, 12 are \emph{over-crediting}  that share one cause: the detector treats any repeated calculator call as a check. It therefore credits verification when a looping model repeats a call, when the same expression appears twice within one reasoning step, and when a model quietly corrects a wrong value after repeating the call without acknowledging the error. The 2 \emph{misses} are the reverse: the model stated that a tool result was wrong and why but did not recompute, which the detector does not count. They are concentrated by model: nemotron-3-ultra contributes ten, gpt-oss-20b three, and claude-opus-4.8 one; the other three models show none (Appendix~\ref{app:verify-disagreement}). \emph{Blind annotators.} Two annotators, blind to all labels, labeled 40 randomly sampled pilot traces and 60 fault events (Appendix~\ref{app:blind}). D-correct agreed on every trace. D-verify again agreed only fairly ($\kappa=0.28$ and $-0.18$), and the annotators agreed with each other no better ($\kappa=0.29$): verification is hard to judge from a trace, whichever traces are chosen. The quiet-replacement flag had 75--87\% precision and 50--71\% recall, so the shares in Table~\ref{tab:audit} are approximate, though their 10--85\% spread across models is far larger.

\section{Discussion}

Labels such as ``verified,'' ``correct,'' or ``fault did not propagate'' are produced by measurement procedures that can miss what happened in the trace. A model may recover without the detector's expected pattern, a loop may be counted as verification, a propagation question may be undecidable, and a serving failure may look like a model failure. The scaled audit shows these are not artifacts of a small sample, and the chemistry task shows they are not specific to arithmetic.
 
The chemistry results sharpen the point about silent override. Where a model can recompute a tool's output, it often corrects the tool quietly; where it cannot, it trusts the tool almost completely. Neither behavior is visible to a detector that only looks for re-execution, and both disappear in an aggregate score. In settings where the tool represents a sensor, laboratory result, or financial record, the original output, the value later used, and the detector label should all remain visible. Our experiments do not test such applications directly, but they show why evaluations should report verification behavior per tool and per task rather than as a single trait of the model.

\subsection{Limitations and Future Work}

This study has the following limitations: First, the detectors audited are rules from our own pipeline; the point is not that these particular rules are weak, but that rule-based trace labels need reported evidence and applicability, and we expect similar gaps in other harnesses. Second, human review covers 90 pilot traces by one reviewer, selected by the reviewer rather than sampled, and used to revise D-impact; the scaled and chemistry results use automatic labels only. Two blind annotators (Appendix~\ref{app:blind}) test this selection; larger samples and human labels for the scaled and chemistry runs remain future work. Fault exposure depends on how often each model calls a tool, so fault effects condition on the first call. Third, the chemistry knowledge base is small (22 molecules), and faulted calculator results there are rounded to four decimals while clean results are not, a cue a model could exploit; any such effect would make the reported trust a lower bound. Finally, the ablation covers two models and two resource controls; it shows that limits change what is observed, not how to choose them.

\section{Conclusion}

%Agent evaluation is itself a measurement process. In our controlled audit, detector outputs could be correct under their implemented rules yet still be incomplete or misleading descriptions of the underlying trace. \textbf{\ref{contrib:C1}} makes that measurement state explicit, \textbf{\ref{contrib:C2}} shows where the gaps appear in practice, and \textbf{\ref{contrib:c3}} makes the supporting evidence inspectable. The goal is not to replace automated evaluation, but to make clear when its outputs can be trusted and when the trace deserves closer review.

Agent evaluation is itself a measurement process. In a controlled audit at scale and across two tasks, detector outputs could be correct under their rules yet incomplete or misleading descriptions of the trace. \textbf{\ref{contrib:C1}} makes the measurement state explicit, \textbf{\ref{contrib:C2}} shows where the gaps appear in practice, and \textbf{\ref{contrib:c3}} makes the supporting evidence inspectable. The goal is not to replace automated evaluation, but to make clear when its outputs can be trusted and when the trace deserves closer review.

\subsection*{AI use statement}
We used Claude Code Agent to help develop the experiment toolkit and analysis scripts and assist with debugging. LLMs served as the evaluated agents in the experiments described in the paper. Grammarly was used to improve grammar and sentence-level flow during manuscript preparation. The authors reviewed all experimental design choices, analyses, numbers, interpretation, and final claims.

\subsection*{Ethics statement}
This work studies synthetic arithmetic and chemistry tasks and does not use private user data or human-subject data. The broader motivation is safety-relevant: agent-evaluation systems should not hide measurement failures when their outputs may inform high-stakes decisions.

\subsection*{Reproducibility statement}
Section~\ref{sec:tasks} and Appendix~\ref{app:repro} give the task, question selection, fault model and rate, seeds, sampling settings, harness limits, detector decision rules, scoring, and error handling; Appendix~\ref{app:chem} gives the chemistry knowledge base and question construction. Every reported table is generated from the trace files by one command, and the fault injector is verified by replaying every recorded tool call.

%%\subsubsection*{Acknowledgments}
\raggedbottom
\bibliographystyle{plainnat}
\bibliography{references}

@inproceedings{bean2025measuring,
  title     = {Measuring What Matters: Construct Validity in Large Language Model Benchmarks},
  author    = {Bean, Andrew M. and Kearns, Ryan Othniel and Romanou, Angelika and Hafner, Franziska Sofia and Mayne, Harry and Batzner, Jan and Foroutan, Negar and Schmitz, Chris and Korgul, Karolina and Batra, Hunar and Deb, Oishi and Beharry, Emma and Emde, Cornelius and Foster, Thomas and Gausen, Anna and Grandury, Mar{\'i}a and Han, Simeng and Hofmann, Valentin and Ibrahim, Lujain and Kim, Hazel and Kirk, Hannah Rose and Lin, Fangru and Liu, Gabrielle Kaili-May and Luettgau, Lennart and Magomere, Jabez and Rystr{\o}m, Jonathan and Sotnikova, Anna and Yang, Yushi and Zhao, Yilun and Bibi, Adel and Bosselut, Antoine and Clark, Ronald and Cohan, Arman and Foerster, Jakob and Gal, Yarin and Hale, Scott A. and Raji, Inioluwa Deborah and Summerfield, Christopher and Torr, Philip H. S. and Ududec, Cozmin and Rocher, Luc and Mahdi, Adam},
  booktitle = {Advances in Neural Information Processing Systems 38, Datasets and Benchmarks Track},
  year      = {2025},
  eprint    = {2511.04703},
  archivePrefix = {arXiv}
}

@article{bhat2026benchmarking,
  title   = {Benchmarking the Benchmarks: A Validity Audit of Tool-Calling Evaluation},
  author  = {Bhat, Vishvesh and Vaghasiya, Jay and Mohsin, Muhammad Ahmed and Aali, Asad},
  journal = {arXiv preprint arXiv:2607.02577},
  year    = {2026}
}

@inproceedings{cheng2025tool,
  title     = {Investigating Tool-Memory Conflicts in Tool-Augmented {LLMs}},
  author    = {Cheng, Jiali and Pan, Rui and Amiri, Hadi},
  booktitle = {R2-FM Workshop at International Conference on Machine Learning (ICML)},
  year      = {2025},
  eprint    = {2601.09760},
  archivePrefix = {arXiv}
}

@inproceedings{gao2026graph,
  title     = {Graph of Trace: Visualizing Execution Traces of Scientific Agents},
  author    = {Gao, Tianci and Li, Haoxuan and Li, Jianhe and Zhao, Tianxiang and Shi, Runze and Wang, Weiran and Wu, Zezhao and Mi, Lu},
  booktitle = {Proceedings of the 64th Annual Meeting of the Association for Computational Linguistics (Volume 3: System Demonstrations)},
  pages     = {297--306},
  year      = {2026},
  publisher = {Association for Computational Linguistics},
  address   = {San Diego, California, United States}
}

@article{gurram2026evaluating,
  title   = {Evaluating Tool-Using Language Agents: Judge Reliability, Propagation Cascades, and Runtime Mitigation in {AgentProp-Bench}},
  author  = {Gurram, Bhaskar},
  journal = {arXiv preprint arXiv:2604.16706},
  year    = {2026}
}

@article{lu2024agentlens,
  title   = {{AgentLens}: Visual Analysis for Agent Behaviors in {LLM}-based Autonomous Systems},
  author  = {Lu, Jiaying and Pan, Bo and Chen, Jieyi and Feng, Yingchaojie and Hu, Jingyuan and Peng, Yuchen and Chen, Wei},
  journal = {arXiv preprint arXiv:2402.08995},
  year    = {2024}
}

@article{varlamov2026memtoc,
  title   = {{MemToC}: Benchmarking Memory-Tool Conflict Resolution in Large Language Models},
  author  = {Varlamov, Arseniy and Zinnatullin, Rishat and Rykov, Elisei and Panchenko, Alexander and Alimova, Ilseyar},
  journal = {arXiv preprint arXiv:2608.26295},
  year    = {2026}
}

@article{yao2026harness,
  title   = {Harness-Bench: Measuring Harness Effects across Models in Realistic Agent Workflows},
  author  = {Yao, Yilun and Tan, Xinyu and Liu, Chao-Hsuan and Li, Yaoming and Wang, Zhengyang and Yu, Wenhan and Tan, Zhewen and Tian, Yuxuan and Zhao, Guangxiang and Sun, Lin and Zhang, Xiangzheng and Yang, Tong},
  journal = {arXiv preprint arXiv:2605.27922},
  year    = {2026}
}

@article{cobbe2021gsm8k,
  title   = {Training Verifiers to Solve Math Word Problems},
  author  = {Cobbe, Karl and Kosaraju, Vineet and Bavarian, Mohammad and Chen, Mark and Jun, Heewoo and Kaiser, Lukasz and Plappert, Matthias and Tworek, Jerry and Hilton, Jacob and Nakano, Reiichiro and Hesse, Christopher and Schulman, John},
  journal = {arXiv preprint arXiv:2110.14168},
  year    = {2021}
}

@misc{wang2026reflect,
  title        = {Time to {REFLECT}: Can We Trust {LLM} Judges for Evidence-based Research Agents?},
  author       = {Wang, Leyao and He, Yanan and Chen, Peng and Yehudai, Asaf
                  and Liu, Yixin and Ying, Rex and Shmueli-Scheuer, Michal
                  and Cohan, Arman},
  year         = {2026},
  eprint       = {2605.19196},
  archivePrefix = {arXiv},
  url          = {https://arxiv.org/abs/2605.19196}
}

@misc{verma2026agentjudgebench,
  title        = {{AgentJudgeBench}: A Multi-Difficulty Benchmark for Evaluating
                  {LLM} Judges on Agentic Tool-Calling},
  author       = {Verma, Abhigya and Saha, Amit Kumar and
                  Subramanian, Seganrasan and Aluru, Sai Harshitha},
  year         = {2026},
  eprint       = {2608.26623},
  archivePrefix = {arXiv},
  url          = {https://arxiv.org/abs/2608.26623}
}

@misc{wu2026clawtrack,
  title        = {{ClawTrack}: Towards Trace-Level Evaluation and Improvement
                  of Real-World Autonomous Agents},
  author       = {Wu, Xingjian and Zhu, Xuhang and Liu, Xingchen and Liu, Junlin
                  and Wang, Jianing and Guo, Linsen and Li, Xiaoyu and
                  Cao, Xuezhi and Cai, Xunliang},
  year         = {2026},
  eprint       = {2607.28037},
  archivePrefix = {arXiv},
  url          = {https://arxiv.org/abs/2607.28037}
}

@misc{milner2026ctfabacus,
  title        = {How Do {LLM} Agents Actually Get the Flag?
                  {T}race-Level Provenance for Agentic Offensive Security Evaluation},
  author       = {Milner, Kimberly and Shao, Minghao and Rani, Nanda and Xi, Haoran
                  and Putrevu, Venkata Sai Charan and Udeshi, Meet
                  and Shukla, Sandeep K. and Krishnamurthy, Prashanth
                  and Khorrami, Farshad and Shafique, Muhammad and Karri, Ramesh},
  year         = {2026},
  eprint       = {2608.26237},
  archivePrefix = {arXiv},
  url          = {https://arxiv.org/abs/2608.26237}
}

@inproceedings{yehudai2026agenticclear,
  title     = {Agentic {CLEAR}: Automating Multi-Level Evaluation of {LLM} Agents},
  author    = {Yehudai, Asaf and Eden, Lilach and Shmueli-Scheuer, Michal},
  booktitle = {Proceedings of the 64th Annual Meeting of the Association for
               Computational Linguistics (Volume 3: System Demonstrations)},
  year      = {2026},
  month     = jul,
  address   = {San Diego, California, United States},
  publisher = {Association for Computational Linguistics},
  pages     = {750--763},
  doi       = {10.18653/v1/2026.acl-demo.74},
  url       = {https://aclanthology.org/2026.acl-demo.74/}
}

@article{pham2025chemgraph,
  title   = {{ChemGraph}: An Agentic Framework for Computational Chemistry Workflows},
  author  = {Pham, Thang D. and Tanikanti, Aditya and Ke{\c{c}}eli, Murat},
  journal = {arXiv preprint arXiv:2506.06363},
  year    = {2025}
}

@misc{chemgraphlogs,
  title        = {{ChemGraph} agent logs},
  author       = {{Autonomous-Scientific-Agents}},
  howpublished = {Hugging Face dataset, \url{https://huggingface.co/datasets/Autonomous-Scientific-Agents/logs}},
  year         = {2026},
  note         = {Accessed October 2026}
}

@article{batatia2023mace,
  title   = {A foundation model for atomistic materials chemistry},
  author  = {Batatia, Ilyes and Benner, Philipp and Chiang, Yuan and Elena, Alin M. and Kov{\'a}cs, D{\'a}vid P. and others},
  journal = {arXiv preprint arXiv:2401.00096},
  year    = {2025}
}

@article{bannwarth2019gfn2,
  title   = {{GFN2-xTB}---An Accurate and Broadly Parametrized Self-Consistent Tight-Binding Quantum Chemical Method with Multipole Electrostatics and Density-Dependent Dispersion Contributions},
  author  = {Bannwarth, Christoph and Ehlert, Sebastian and Grimme, Stefan},
  journal = {Journal of Chemical Theory and Computation},
  volume  = {15},
  number  = {3},
  pages   = {1652--1671},
  year    = {2019},
  doi     = {10.1021/acs.jctc.8b01176}
}

@inproceedings{sun2024toolsfail,
  title     = {Tools Fail: Detecting Silent Errors in Faulty Tools},
  author    = {Sun, Jimin and Min, So Yeon and Chang, Yingshan and Bisk, Yonatan},
  booktitle = {Proceedings of the 2024 Conference on Empirical Methods in Natural Language Processing},
  pages     = {14272--14289},
  year      = {2024},
  publisher = {Association for Computational Linguistics},
  doi       = {10.18653/v1/2024.emnlp-main.790}
}

@inproceedings{wu2024clasheval,
  title     = {{ClashEval}: Quantifying the tug-of-war between an {LLM}'s internal prior and external evidence},
  author    = {Wu, Kevin and Wu, Eric and Zou, James},
  booktitle = {Advances in Neural Information Processing Systems (Datasets and Benchmarks Track)},
  year      = {2024}
}

@inproceedings{xie2024chameleon,
  title     = {Adaptive Chameleon or Stubborn Sloth: Revealing the Behavior of Large Language Models in Knowledge Conflicts},
  author    = {Xie, Jian and Zhang, Kai and Chen, Jiangjie and Lou, Renze and Su, Yu},
  booktitle = {International Conference on Learning Representations},
  year      = {2024}
}

@article{zhang2026,
  title   = {Don't Blindly Trust It: How Unreliable Feedback Breaks Tool-Using {LLM} Agents},
  author  = {Zhang, Chubin and Wan, Zhenglin and Yu, Xingrui and Zhou, Pengfei and Zhao, Wangbo and Wu, Jingxuan and Zhou, Yaxin and Tsang, Ivor},
  journal = {arXiv preprint arXiv:2606.21409},
  year    = {2026}
}

@inproceedings{mittal2026,
  title     = {When Tool Outputs Go Wrong: Belief Revision in Tool-Using Agents},
  author    = {Mittal, Avni},
  booktitle = {ICML 2026 Workshop on Failure Modes in Agentic AI: Reproducible Triggers, Trace Diagnostics, and Verified Fixes},
  year      = {2026},
  url       = {https://openreview.net/pdf?id=IFPcrT092O}
}

@article{tao2026,
  title   = {When Tools Silently Lie: Evaluating and Mitigating Blind Compliance in Tool-Augmented Data Agents},
  author  = {Tao, Zifu and Yin, Changqing},
  journal = {arXiv preprint arXiv:2609.37153},
  year    = {2026}
}

@inproceedings{kaynar2026,
  title     = {When Tools Fail, Agents Fabricate: Silent Failure and Verification-First Oversight in Tool-Using {LLM} Agents},
  author    = {Kaynar, G{\"u}n and Yan, Boshen and Kingsford, Carl},
  booktitle = {NeurIPS 2026 Workshop: Who Verifies the Agents? Toward Reliable Agent Development},
  year      = {2026}
}

\appendix
%%\section{Additional audit results}
%%\label{app:audit}
\section{Reproducibility details}
\label{app:repro}
\begin{table}[H]
\centering
\caption{Settings for all runs.}
\label{tab:settings}
\small
\begin{tabularx}{\linewidth}{@{}l Y@{}}
\toprule
Questions & GSM8K training split in its published order: problems 0--29 (pilot) and 0--499 (scaled); identical for every model. Chemistry: 280 questions, identical for every model. \\
Fault model & Per-call probability 0.15. Integers: true value $\pm1$, $\pm2$, or an adjacent-digit transposition (uniform over distinct candidates). Non-integers: $\pm5\%$, rounded to four decimals. Chemistry retrieval: neighbouring method/temperature, closest-magnitude molecule, or a 0.5--3\% perturbation; search: a molecule sharing most of the true molecule's records. \\
Seeds & Per-question stream seeded with $\mathrm{sha256}(\textit{seed}{:}\textit{index})$. Pilot: seed 50. Ablation: seeds 50, 51 and 52. Scaled and chemistry: seeds 50 and 51; chemistry clean control: seed 50, no faults. On the 30 pilot problems, the first calculator call is faulted in exactly the same traces in the pilot and the scaled seed-50 run (30/30 for claude-opus-4.8). \\
Sampling & Temperature 0 (scaled claude-sonnet-5: provider default; the gateway rejects the parameter); 1,024 output tokens per turn; 50 steps (pilot Claude loop: 15 steps, never reached). \\
Scoring & Number after the last \texttt{FINAL ANSWER:}; exact match for GSM8K, with a scale word immediately after the number (``192 million'') accepted for a matching gold value (one trace in the scaled audit); chemistry tolerance $\max(0.02,\,0.006n)$\,eV for an answer combining $n$ values. \\
Errors & Serving errors (HTTP errors, tool calls the endpoint returned as text) are excluded from denominators and reported as coverage; step-limit hits and stops without an answer are scored as not correct and reported as censored. \\
Duplicates & Where a question ran twice under identical conditions, we kept a completed run over an errored one, otherwise the earlier run, independent of its outcome. \\
Llama & Llama-3.1's chat template rejects a history turn with several tool calls; for Llama only, such turns are re-sent as consecutive single-call turns with unchanged calls, order, and outputs. \\
\bottomrule
\end{tabularx}
\end{table}
 
\section{Statistical details}
\label{app:stats}
First-call effects with Newcombe 95\% intervals for the difference of two proportions: claude-opus-4.8 6.6 [1.9, 13.1], claude-sonnet-5 7.3 [2.3, 14.0], Gemma 5.2 [0.9, 11.3], gpt-oss-120b 8.7 [3.0, 16.0], gpt-oss-20b 9.0 [3.0, 16.5], Mistral 12.9 [4.8, 22.9], nemotron-3-ultra 8.4 [1.9, 16.3], Llama 53.9 [45.7, 60.7]. These treat the two seeds of a question as independent. Table~\ref{tab:bootstrap} instead resamples questions rather than traces, so the two seeds of a question move together, and counts fault outcomes per trace so that looping traces with many events do not dominate.
 
\begin{table}[H]
\centering\small
\caption{Question-clustered bootstrap estimates (2000 resamples of questions; both seeds of a question are resampled together). Fault effect: accuracy when the first tool call is clean minus when it is faulted, in points. Used (per trace): share of traces with at least one decidable fault event in which a bad value was used; the last column excludes censored traces and traces with more than 20 tool calls (looping).}
\label{tab:bootstrap}
\resizebox{\linewidth}{!}{%
\begin{tabular}{@{}lrrrrrr@{}}
\toprule
Model & Traces & Accuracy (\%) & Fault effect (pts) & Used, per event & Used, per trace (\%) & excl.\ looping (\%) \\
\midrule
Mistral-Large-3-675B & 660 & 87.0 [83.8, 90.1] & 12.9 [3.1, 23.2] & 23/237 & 11.2 [6.9, 15.6] & 11.2 [6.9, 15.6] \\
claude-opus-4.8 & 1000 & 93.7 [92.0, 95.5] & 6.6 [0.8, 13.1] & 43/460 & 10.9 [7.8, 13.9] & 10.9 [7.8, 13.9] \\
claude-sonnet-5$^\ast$ & 1000 & 92.9 [91.0, 94.7] & 7.3 [1.2, 13.4] & 42/385 & 14.4 [10.3, 19.0] & 14.4 [10.3, 19.0] \\
Gemma-4-31B & 1000 & 94.5 [92.6, 96.2] & 5.2 [-0.1, 11.0] & 16/602 & 3.9 [2.2, 5.8] & 3.8 [2.0, 5.7] \\
Llama-3.1-70B & 1000 & 69.9 [66.5, 73.7] & 53.9 [45.7, 61.9] & 100/159 & 63.2 [55.3, 71.2] & 63.2 [55.3, 71.2] \\
nemotron-3-ultra & 1000 & 83.1 [80.2, 85.9] & 8.4 [1.1, 16.0] & 23/1006 & 6.4 [3.9, 9.4] & 8.3 [5.0, 12.4] \\
gpt-oss-120b & 1000 & 90.2 [87.7, 92.4] & 8.7 [1.7, 15.7] & 31/387 & 11.7 [7.9, 15.6] & 12.0 [8.1, 16.0] \\
gpt-oss-20b & 1000 & 87.4 [84.9, 90.0] & 9.0 [2.4, 15.8] & 14/317 & 7.6 [4.1, 11.6] & 7.9 [4.1, 11.9] \\
\bottomrule
\end{tabular}}
\end{table} 
 
\section{Chemistry task details}
\label{app:chem}
Generated questions: 57 single lookups, 150 differences between two molecules (sampled from 238), 10 differences between methods for one molecule, 20 Gibbs-energy changes between temperatures, and 16 reaction energies over 14 balanced reactions (products minus reactants with coefficients). The 27 original ChemGraph questions are included unchanged. Tolerance grows with the number of retrieved values so that rounding each to three decimals is still scored correct, while every injected retrieval fault falls outside it. Llama omitted the answer line on 25 of its 840 chemistry traces (17 contained the correct value in prose); these are scored as unanswered, as in arithmetic. The knowledge-base fingerprint is stored in every trace, and replay refuses traces produced with a different knowledge base.

\section{Seed-50 ablation detail}
\label{app:ablation}
\begin{table}[H]
\centering
\caption{Seed-50 ablation cells in detail (the original run). Not compl.: runs stopped by the step limit or without an answer (serving errors are excluded). Faults: faulted / total calculator calls. Prop.: decidable fault events in which the faulted value was used. Fault counts are not comparable across cells because longer runs make more calls.}
\label{tab:ablation-detail}
\small
\begin{tabular}{@{}l l r r r r r r@{}}
\toprule
\textbf{Model} & \textbf{Cell} & \textbf{Acc} & \textbf{Not compl.} & \textbf{Faults} & \textbf{Prop.} & \textbf{Verify} & \textbf{Quiet} \\
\midrule
\multirow{6}{*}{nemotron-3-ultra}
 & steps=25,  tokens=1024 & 23/30 & 7 & 131/874   & 1/91 & 11/30  & 16/30 \\
 & steps=25,  tokens=4096 & 27/30 & 3 & 496/3524  & 0/460 & 9/30  & 21/30 \\
 & steps=50,  tokens=1024 & 21/30 & 9 & 174/1148  & 0/119  & 12/30 & 15/30 \\
 & steps=50,  tokens=4096 & 27/30 & 3 & 518/3627  & 0/459 & 6/30  & 24/30 \\
 & steps=150, tokens=1024 & 24/30 & 6 & 237/1635  & 2/128 & 10/30 & 17/30 \\
 & steps=150, tokens=4096 & 25/30 & 5 & 612/4401  & 0/467 & 9/30  & 21/30 \\
\midrule
\multirow{6}{*}{gpt-oss-120b}
 & steps=25,  tokens=1024 & 27/30 & 1 & 16/89  & 2/9 & 6/30 & 1/30 \\
 & steps=25,  tokens=4096 & 29/30 & 0 & 9/63  & 1/8  & 7/30 & 1/30 \\
 & steps=50,  tokens=1024 & 27/30 & 0 & 9/66   & 2/8 & 7/30 & 0/30 \\
 & steps=50,  tokens=4096 & 28/30 & 0 & 9/67   & 1/7  & 4/30 & 3/30 \\
 & steps=150, tokens=1024 & 28/30 & 0 & 11/76   & 1/9 & 8/30 & 0/30 \\
 & steps=150, tokens=4096 & 27/29\textsuperscript{$\ddagger$} & 0 & 11/75  & 1/10  & 4/30 & 3/30 \\
\bottomrule
\end{tabular}
 
\vspace{2pt}
{\footnotesize $^\ddagger$One serving-error trace excluded from the denominator, as in all scaled results.}
\end{table}
 
Event counts such as 0/460 come from at most 30 traces, so they are not 460 independent observations: nemotron's high fault totals reflect looping traces that call the calculator over a hundred times. Appendix~\ref{app:stats} reports fault outcomes per trace and without looping traces.

\subsection{Per-model verification disagreement}
\label{app:verify-disagreement}

Table~\ref{tab:per-model-reliability} reports the per-model distribution
of D-verify disagreements among reviewed traces with a comparable
verification label.

\begin{table}[H]
\centering
\caption{D-verify disagreements among reviewed traces with a comparable
verification label. These counts describe the reviewed cases and should
not be interpreted as population disagreement rates.}
\label{tab:per-model-reliability}

\small
\begin{tabularx}{\linewidth}{
    @{}
    X
    >{\centering\arraybackslash}p{0.25\linewidth}
    @{}
}
\toprule
\textbf{Model} &
\textbf{Disagreements / reviewed} \\
\midrule
nemotron-3-ultra     & 10 / 17 \\
gpt-oss-20b          & 3 / 14  \\
claude-opus-4.8      & 1 / 10  \\
gpt-oss-120b         & 0 / 13  \\
claude-sonnet-5      & 0 / 10  \\
Mistral-Large-3-675B & 0 / 17  \\
\bottomrule
\end{tabularx}
\end{table}

\FloatBarrier
\section{Inspection interfaces}
\label{app:viewer}
\begin{figure}[H]
\centering
\includegraphics[width=\linewidth]{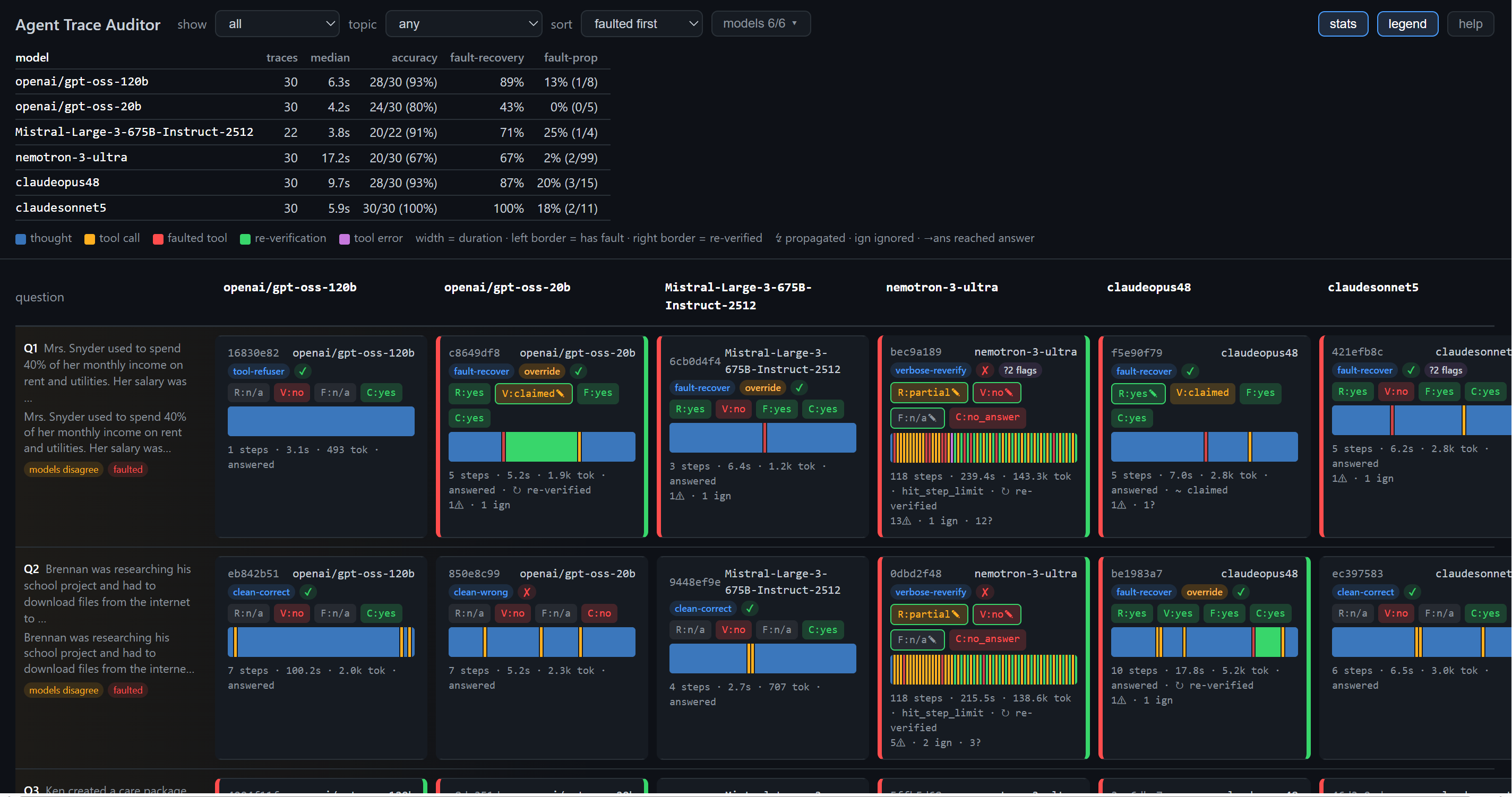}
\caption{\textbf{Corpus view.} Traces aligned by question and model, showing outcome, faults, verification, recovery, and risk flags across runs.}
\label{fig:viewer-corpus}
\end{figure}
 
\begin{figure}[H]
\centering
\includegraphics[width=0.8\linewidth]{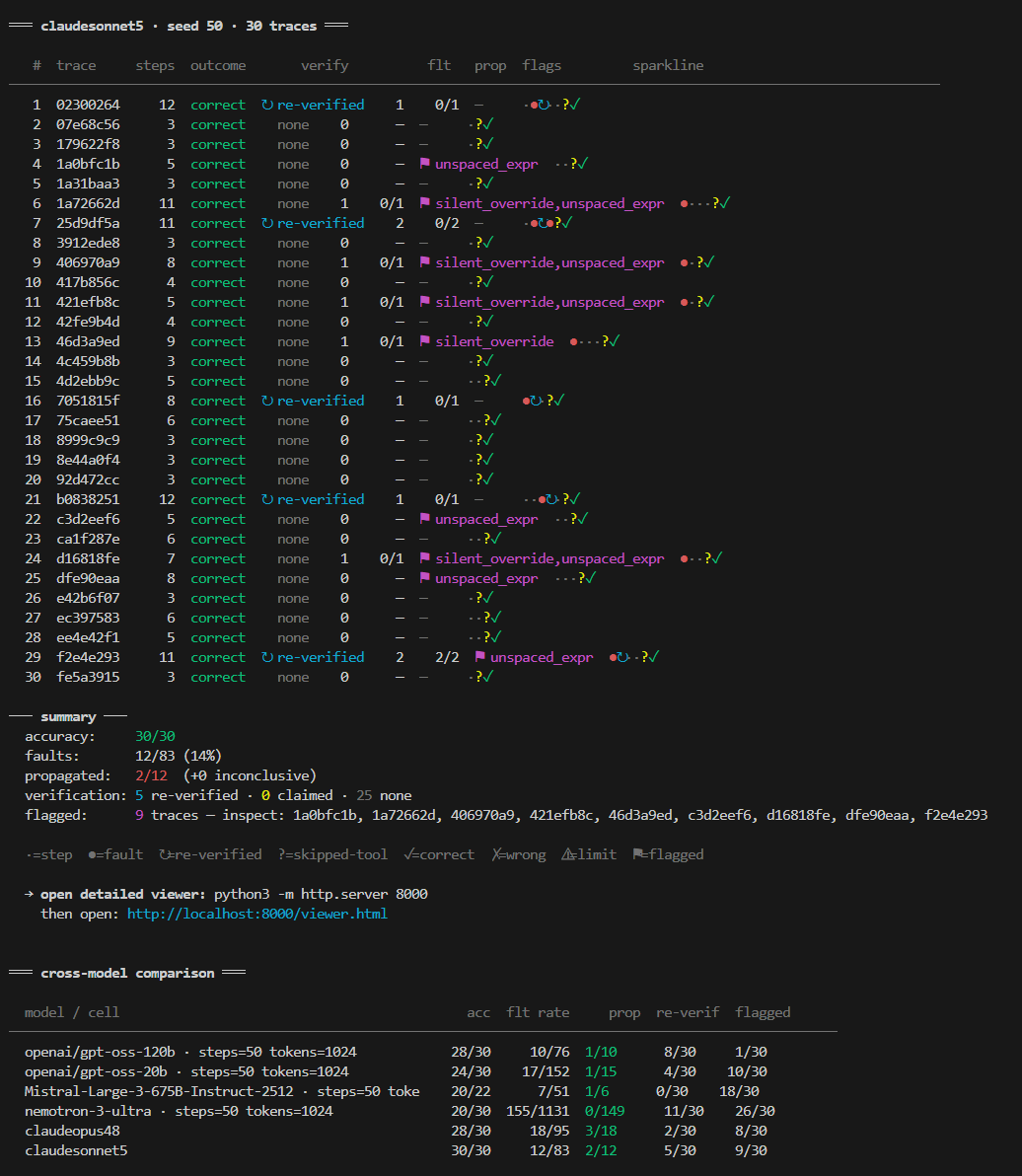}
\caption{\textbf{Terminal summary for rapid triage.} Per-trace outcomes, fault exposure, verification events, and flagged cases before detailed inspection.}
\label{fig:terminal-appendix}
\end{figure}

\FloatBarrier
\section{Blind annotator study}
\label{app:blind}
Two annotators labeled two packets without access to the viewer, the automatic labels, or the first reviewer's labels. The \emph{trace packet} holds 40 pilot traces drawn at random (seed 2027) from all non-empty pilot traces, each shown as text with the question, gold answer, every step, and faulted tool calls marked with the true value; annotators answered the four questions of Table~\ref{tab:agreement}. The \emph{event packet} holds 60 fault events from the arithmetic runs of the seven models other than claude-sonnet-5, whose scaled run was completed after sampling: 30 drawn from the 1,020 events the detector flags as quiet replacements and 30 from the 2,148 other decidable events, shuffled, with flags hidden; for each, annotators judged whether the model later used the true value without re-running the call or saying the tool was wrong, and which value it carried forward. Recall is estimated by weighting each stratum by its population.

\begin{table}[H]
\centering
\caption{Blind annotators (A1, A2). \textbf{Top:} trace packet, agreement on yes/no labels and Cohen's $\kappa$; ``--'' marks $\kappa$ where nearly all labels share one value, so $\kappa$ is undefined or uninformative. \textbf{Bottom:} event packet, precision (exact 95\% interval) and estimated recall of the quiet-replacement flag against each annotator. A1's recall is 83\% when the 8 events with an identical re-run, which the definition excludes, are not counted as misses.}
\label{tab:blind}
\small
\begin{tabular}{@{}lccc@{}}
\toprule
\textbf{Trace packet ($n=40$)} & \textbf{A1 vs detector} & \textbf{A2 vs detector} & \textbf{A1 vs A2} \\
\midrule
D-correct           & 40/40, $\kappa=1.00$ & 40/40, $\kappa=1.00$ & 40/40, $\kappa=1.00$ \\
D-impact / recovery & 17/20, $\kappa=0.32$ & 17/18, $\kappa=0.64$ & 20/20, $\kappa=1.00$ \\
Fault relevance     & 18/21, -- & 21/21, -- & 19/22, -- \\
D-verify            & 31/40, $\kappa=0.28$ & 27/40, $\kappa=-0.18$ & 30/40, $\kappa=0.29$ \\
\bottomrule
\end{tabular}

\vspace{8pt}
\begin{tabular}{@{}lccc@{}}
\toprule
\textbf{Event packet ($n=60$)} & \textbf{Decidable labels} & \textbf{Precision of flag} & \textbf{Estimated recall} \\
\midrule
A1 & 47 & 20/23 = 87\% [66, 97] & 50\% \\
A2 & 55 & 21/28 = 75\% [55, 89] & 71\% \\
\bottomrule
\end{tabular}
\end{table}

On the event packet, the two annotators agreed on which value the model carried forward in 32 of 35 events, but on whether the event was a quiet replacement in only 24 of 42 ($\kappa=0.17$).

The disagreements with the flag mark the boundaries of its definition rather than random errors. In 8 of A1's 10 missed events the model re-ran the identical call, which the definition excludes; A2 labeled all eight as not quiet, and excluding them A1's estimated recall is 83\%. A1's remaining disagreements are a notice phrase such as ``verify'' without an admitted error, a recomputation of the same quantity with a different expression (which the detector does not count as a re-run), a faulted value that disappears from a trace that never answers, and events whose surrounding steps contain no text, where the annotators split or answered ``unclear''.

\end{document}